\documentclass[letterpaper, 10 pt, conference]{ieeeconf}  
\let\labelindent\relax

\IEEEoverridecommandlockouts                              

\usepackage{amsmath}
\usepackage{mathrsfs}
\usepackage{multicol}
\usepackage{multirow}
\usepackage{float}
\usepackage{algorithm}
\usepackage{algpseudocode}
\usepackage[normalem]{ulem}
\usepackage{caption}
\usepackage{subcaption}
\usepackage{setspace}
\usepackage{doi}
\usepackage{color}
\usepackage{cleveref}
\usepackage{enumitem}
\usepackage{graphicx}
\usepackage{url}
\usepackage{tablefootnote}
\usepackage{booktabs}

\title{\LARGE \bf
EmbodiedAgent: A Scalable Hierarchical Approach to Overcome Practical Challenge in Multi-Robot Control
}

\author{Hanwen Wan$^{1, 2}$, Yifei Chen$^{1}$, Yixuan Deng$^{1, 2}$, Zeyu Wei$^{3}$, Dongrui Li$^{4}$, Zexin Lin$^{1, 2}$, \\Donghao Wu$^{1, 2}$, Jiu Cheng$^{1, 2}$, and Xiaoqiang Ji$^{1, 2, \dagger}$
\thanks{$^{*}$This work was partially supported by National Natural Science Foundation of China (Grant No. 62441619), Guangdong Basic and Applied Basic Research Foundation (Grant No. 2022A1515110411, Grant No. 2023A1515012883, and Grant No. 2024A1515240009), Shenzhen Science and Technology Program (Grant No. JCYJ20240813113604006, Grant No. KJZD20240903095730039), and Guangxi Science and Technology Planning Project (Grant No. AA23062031-2, Grant No. AA23062073-2).}%
\thanks{$^{1}$School of Science and Engineering, The Chinese University of Hong Kong, Shenzhen, China.}%
\thanks{$^{2}$Shenzhen Institute of Artificial Intelligence and Robotics for Society, China.}%
\thanks{$^{3}$The School of Computer Science, The University of Sydney, Australia.}%
\thanks{$^{4}$Faculty of Computer and Mathematical Sciences, The Hong Kong Polytechnic University, China.}%
\thanks{$^{\dagger}$The corresponding author is Xiaoqiang Ji whose e-mail is {\tt\small jixiaoqiang@cuhk.edu.cn}}%
}

\begin{document}

\maketitle
\thispagestyle{empty}
\pagestyle{empty}

\begin{abstract}
This paper introduces EmbodiedAgent, a hierarchical framework for heterogeneous multi-robot control. EmbodiedAgent addresses critical limitations of hallucination in impractical tasks. Our approach integrates a next-action prediction paradigm with a structured memory system to decompose tasks into executable robot skills while dynamically validating actions against environmental constraints. We present MultiPlan+, a dataset of more than 18,000 annotated planning instances spanning 100 scenarios, including a subset of impractical cases to mitigate hallucination. To evaluate performance, we propose the Robot Planning Assessment Schema (RPAS), combining automated metrics with LLM-aided expert grading. Experiments demonstrate EmbodiedAgent’s superiority over state-of-the-art models, achieving 71.85\% RPAS score. Real-world validation in an office service task highlights its ability to coordinate heterogeneous robots for long-horizon objectives. 

\end{abstract}

\section{INTRODUCTION}
Heterogeneous multi-robot systems, which leverage diverse robotic capabilities and collaborative synergies, outperform single-robot platforms in complex tasks. However, they require robust planning to coordinate task allocation and ensure consensus among robots \cite{consensus}. Traditional consensus algorithms, constrained by rigid rules, limited adaptability, and scalability challenges, struggle to manage the dynamic complexities of such diverse teams \cite{c2}. On the other hand, studies have investigated the use of machine learning techniques to solve the Hamilton–Jacobi–Bellman (HJB) equation for consensus \cite{ji}. In contrast, Large Language Model (LLM) based agents powered by high-level reasoning introduce a paradigm shift by enabling operators to specify broad objectives through unstructured commands \cite{multiplan, a5, a6, a7}. Within hierarchical embodied intelligence systems, these intelligent planners decompose missions into fundamental skills and facilitate downstream action policies for grounded execution in dynamic, unstructured real-world environments.

\begin{figure}
    \centering
    \includegraphics[width=1\linewidth]{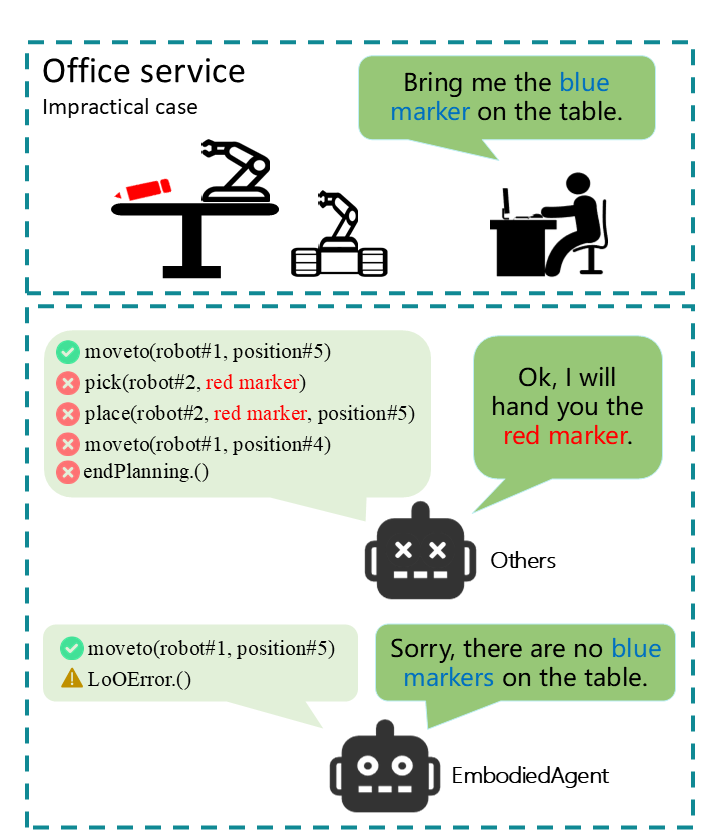}
    \caption{An illustration of a practically infeasible scenario.}
    \label{fig:intro}
\end{figure}

Recent advancements in LLM-driven planners have significantly enhanced the coordination of heterogeneous multi-robot systems. These methods focus on converting mission objectives and operational contexts into structured descriptive frameworks using language-only or language-vision modalities \cite{a1, a2, a3, a4}. However, the majority of these frameworks rely on integrated simulators and have not been extensively validated in real-world environments. For instance, PARTNER \cite{c3} employs a data flywheel in Habitat \cite{c4} for data generation and utilizes a ReAct \cite{react} agent for action planning but overlooks impractical scenarios. Smart-LLM \cite{c9} operates within a few-shot prompting paradigm, decomposing the planning process into task decomposition, coalition formation, and task allocation. This approach is validated in the AI2-THOR simulation environment. Similarly, COHERENT \cite{c10} employs a comparable multi-stage process to decompose complex tasks, integrating self-reflection feedback for task correction. In contrast, EMMA \cite{c11} does not directly infer using vision-language models (VLMs), but instead utilizes GPT-4 as a "teacher" to train embodied agents through interactive cross-modal imitation learning.

In order to improve robustness and mitigate hallucination of LLMs, prior works explore methods like fine-tuning with robot planning specific data \cite{multiplan} and adding a feedback loop \cite{ada}. Specifically, MultiPlan \cite{multiplan} fine-tunes a language model to enable robust task allocation and consensus among heterogeneous robotic platforms, showcasing few-shot transfer capabilities for novel tasks and robots without additional training. ProgPrompt \cite{c5} improves human-robot interaction by incorporating real-time sensory data and environmental constraints. Meanwhile, planners designed for multi-robot systems tackle challenges in task decomposition, allocation, and execution across simulation and real-world settings. TREE-PLANNER \cite{c6} operates within a fixed action space, aggregates sampled plans into an action tree, and refines the planning sequence in subsequent modules, with validation conducted in VirtualHome \cite{c7}. 

However, these planning fail when tasked with impractical or counterfactual scenarios \cite{c8}. Impractical error occur when a planner detects that the required action violates physical constraints or operational limitations inherent to the multi-robot system. As illustrated in \cref{fig:intro}, the user asks for a blue marker while there are no blue markers on the table. These errors can lead to hallucinated plans or counterfactual execution, which are particularly critical in real-world applications, both industrial and domestic. Additionally, these systems exhibit limited adaptability when extrapolated to novel task configurations, undermining their effectiveness in cross-domain applications. These challenges highlight the necessity for a robust and generalized large language model (LLM) that bridges theoretical task abstraction with practical feasibility, while effectively managing diverse environments and operational constraints.

In response to these limitations, this work introduces a hierarchical \textbf{Embodied} system with an \textbf{Agent}-based planner, named \textbf{EmbodiedAgent}. EmbodiedAgent leverages a next-action prediction paradigm to establish a heterogeneous multi-robot control system. The core agent generates a single action and its corresponding arguments per inference, terminating upon receiving an end-of-planning signal, thus ensuring a controlled and concise execution process. To address the aforementioned challenges, we enhance the planner’s robustness and generalizability through supervised fine-tuning. Extended from previous work MultiPlan \cite{multiplan}, we present MultiPlan+, a large-scale dataset comprising 100 scenarios with over 18,000 tasks, enriched with a subset of impractical cases to mitigate hallucinations. Additionally, we develop an agent based on a fine-tuned language model equipped with function calling capabilities and structured memory. Specifically, robot skills, termination signals, and error signals related to impractical cases are encapsulated as tools, while planning history is organized within the structured memory. For low-level execution, we employ specialized policies trained on individual basic tasks to ensure reliable and robust performance. Furthermore, we propose a comprehensive \textbf{R}obot \textbf{P}lanning \textbf{A}ssessment \textbf{S}chema (\textbf{RPAS}), which moves beyond error-type diagnostics to emphasize stratified success rates assessed through both human evaluation and automated grading. Code and dataset are open-sourced \footnote{\url{https://github.com/HaronW/EmbodiedAgent}}. In summary, our main contributions are as follows:

\begin{enumerate}
    \item Propose EmbodiedAgent, a hierarchical embodied heterogeneous multi-robot control system. Leveraging agent-based techniques, EmbodiedAgent enables robust coordination of diverse robotic platforms to accomplish complex, long-horizon tasks.
    \item Introduce MultiPlan+, a large-scale dataset with more than 18000 tasks from 100 indoor and outdoor scenarios in the format of next-action prediction. MultiPlan+ is augmented with a subset of impractical cases, aiming at addressing the practical infeasible problem.
    \item Develop RPAS, a systematic evaluation framework for quantifying performance in embodied AI systems, with a focus on planning robustness and task success rates.
    \item Validate the system through comparison experiment and real-world deployment across heterogeneous robot teams, demonstrating scalability and effectiveness in unstructured environments.
\end{enumerate}

\section{PROBLEM FORMULATION}
This work presents a unified and scalable framework for deriving optimal action plans using agents in complex operational environments. Building on the MultiPlan \cite{multiplan}, we employ an indexed positions system to represent real-world points, thereby reducing ambiguities and improving spatial accuracy. The framework formalizes a multi-robot planning task through three interconnected components: the mission description $\mathbf{M} = \{\text{scenario}, \text{task}\}$, the environment configuration $\mathbf{E} = \{\text{workspace}, \text{robot}, \text{object}, \text{user}\}$, and the planning memory $\mathbf{P}_t = \{p_1, p_2, \ldots, p_t\}$. $\mathbf{M}$ encapsulates the overall goal, where the \textit{scenario} defines the working environment and the \textit{Task} specifies the precise objective to be planned. \textit{Robot} in the environment configuration registers available robots, their defined skills, and workload limits. The \textit{User} entry defines humans interacting with the multi-robot system. Meanwhile, the planning memory maintains a short-term record of previously executed actions, thereby providing the contextual grounding necessary for informed sequential decision-making.

\begin{figure*}
    \centering
    \includegraphics[width=0.97\linewidth]{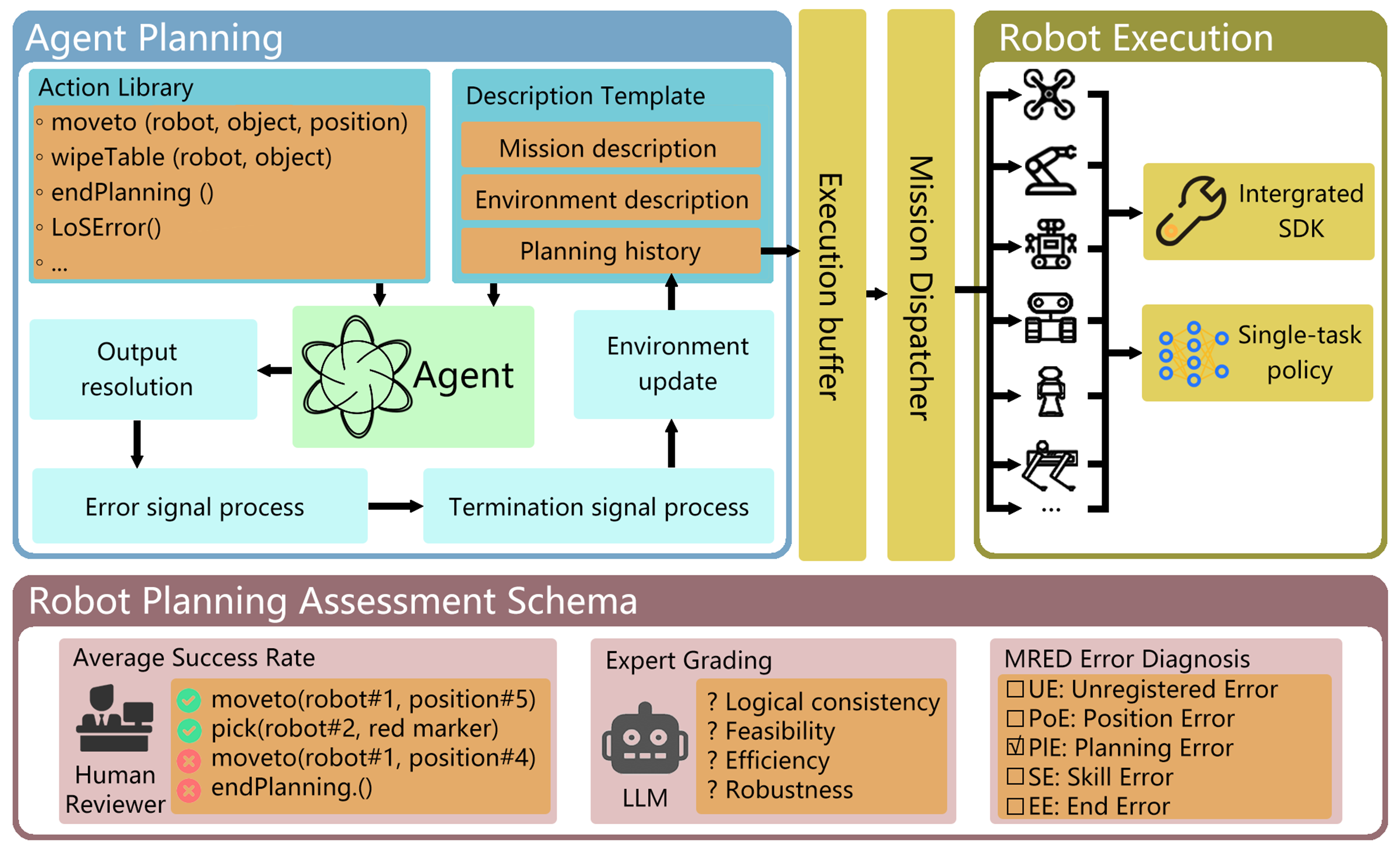}
    \caption{\textbf{The pipeline of EmbodiedAgent} The "Agent Planning" section includes an action library, which defines functions, along with a description template that manages mission, environment, and planning history. The agent's decision-making process involves output resolution, error signal processing, and termination signal processing. Mission dispatcher handles the execution of tasks, interacting with various robot types via an integrated SDK and single-task policies. The RPAS evaluation framework provides a framework for evaluating task performance, including the Average Success Rate, expert grading criteria, and error diagnosis using the MRED system.}
    \label{fig:pipeline}
\end{figure*}

Given the planning state $s_t = \{\mathbf{M}, \mathbf{E}, \mathbf{P}_t\}$ at planning step $t$, the objective of the agent-based planner $\pi$ is to generate the next planning $p_{t+1}$. The next action prediction planning loop of EmbodiedAgent ends with the termination signal $\epsilon_{end}$ or throwing impractical errors, formulated as: $\mathcal{E} = \{\epsilon_{LoA}, \epsilon_{LoO}, \epsilon_{LoS}, \epsilon_{LoL}\}$. Here, $\epsilon_{end}$ signifies successful task completion, while the other signals denote various execution failures. The complete planning objective can be formalized as finding an optimal action sequence that successfully accomplishes the mission while satisfying all operational constraints:

\begin{equation}
    \begin{gathered}
        p_{t+1} = \mathop{\pi}(\mathbf{P}_t \mid \mathbf{M}, \mathbf{E}), \\
        p_{t+1} \in \{ a, \epsilon_{end}, \mathcal{E} \},
    \end{gathered}
    \label{eq:next_action}
\end{equation}
where $a$ is an atomic robot skill that is systematically encapsulated as a callable tool within the planning system.

In summary, we have presented a systemic framework for multi-robot planning that combines mission specifications, environmental configurations, and planning memory. The system generates action sequences through an iterative next-action prediction approach while respecting operational constraints. This formulation provides a foundation for developing robust planning strategies in real-world scenarios.

\section{METHODS}
\subsection{EmbodiedAgent Architecture} 
The EmbodiedAgent is designed as a hierarchical control framework for heterogeneous multi-robot systems, enabling task decomposition through the dynamic composition of diverse robot skills. As illustrated in \cref{fig:pipeline}, the architecture comprises two primary layers: a high-level planner and a low-level execution module, connected by execution buffer and mission dispatcher. The process of EmbodiedAgent is provided in \cref{alg:ea}.

\begin{algorithm}
\caption{EmbodiedAgent}
\label{alg:ea}
\begin{algorithmic}[1]
\State \textbf{Input}: Environment description $D$ (JSON format)
\State \textbf{Output}: Updated description with action sequence

\State \textbf{Initialize} robot states $R$, action history $H$

\While{step $<$ threshold \textbf{and not} termination}
    \State $Q \gets D \oplus R$ 
    \State$res \gets \pi(Q)$ \Comment{get LLM response}
    \State  function $f$, arguments $a \gets \text{extract}(res)$
    
    \If{$f$ is termination (endPlanning)}
        \State Log termination in $H$
        \State \textbf{break} loop
    \ElsIf{$f$ is error signal (Lo*Error)}
        \State Log error type in $H$
        \State \textbf{break} loop
    \Else
        \State Verify $f \in$ valid skills list
        \State Execute $f$ with arguments $a$
        \If{execution succeeds}
            \State Update environment $D \gets \text{new state}$
            \State Update robot states $R \gets \text{new configuration}$
            \State Log action $f(a)$ in $H$
        \Else
            \State Handle argument/environment mismatch
            \State Record error in $H$
            \State \textbf{break} loop
        \EndIf
    \EndIf
    
\EndWhile

\State Save $\{D, H\}$

\end{algorithmic}
\end{algorithm}

\textbf{The high-level planner} consists of an LLM-based reasoning engine, a tool library, and an embedded memory module. The planner interprets task goals and generates sequential actions by invoking appropriate tools from the library, where each tool abstracts a robot skill with standardized inputs and outputs. The tool library includes functions for task completion signals and error interrupts, ensuring robust task decomposition across diverse robots. To maintain contextual awareness, a memory module stores structured records of prior actions and environmental states, iteratively updating prompts for subsequent planning. An environment interface validates generated actions against constraints like collision avoidance and dispatches verified commands to robots.

\textbf{The low-level execution module} translates high-level plans into actionable instructions for robots via specialized SDKs or single-task policies. These policies, trained using reinforcement learning or imitation learning, are tailored for precise operations such as locomotion or object manipulation. High-level actions generated by the planner are executed by invoking corresponding low-level policies, ensuring real-time adaptability to sensor feedback and dynamic environments. This integration of high-level abstraction and low-level precision makes EmbodiedAgent a scalable and versatile solution for heterogeneous multi-robot systems.

\subsection{MultiPlan+ Dataset}
\begin{figure}[t]
    \centering
    \includegraphics[width=1\linewidth]{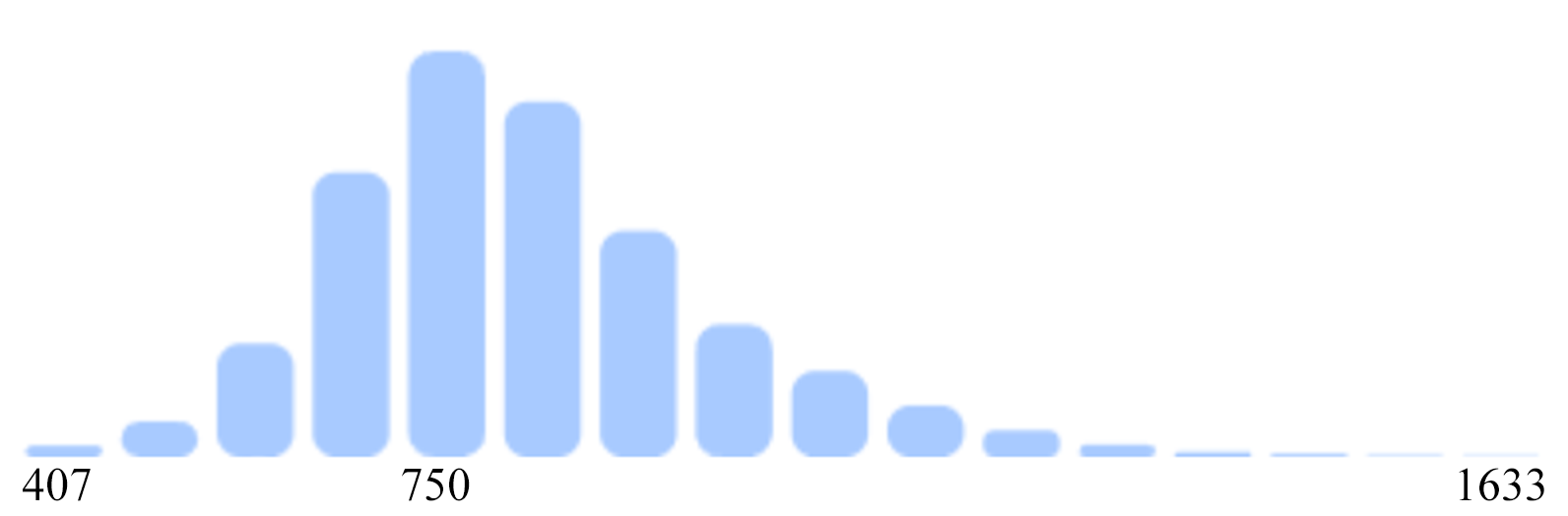}
    \caption{The length of data samples in MultiPlan+ dataset in bytes. Ranging from 406 to 1633.}
    \label{fig:dis}
\end{figure}

MultiPlan+ is a large-scale dataset for heterogeneous multi-robot planning, extended from MultiPlan. The dataset consists of over 18,000 data entries derived from more than 3,400 unique tasks within 100 diverse scenarios, including office, domestic, urban street area, exploratory environments, and etc. The distribution of the data samples are shown in \cref{fig:dis}. Each data entry is annotated with task specification, environmental states, planning memory, and ground-truth action responses. The environment description details the workspace with position points, robot team configurations, objects, and involved users. Different from the MultiPlan dataset, MultiPlan+ adopts the next action prediction paradigm and has been reviewed for diversity.

This dataset is specifically designed for supervised fine-tuning tasks, with annotations structured as callable tools for the planner. These annotations facilitate the training of planners by providing clear, structured feedback on robot actions and planning processes. The data is organized in a role-content format, where each entry includes a role and corresponding content. Valid responses from the assistant are categorized into three types: robot skills, end planning signals, and impractical error signals. 

1. \textbf{Robot skills} represent fundamental capabilities required for task execution, such as locomotion (e.g., moving to a specific location) and manipulation (e.g., picking up objects). Necessary arguments are determined by the planner, ensuring flexibility and adaptability to various scenarios.

2. \textbf{End planning signal} indicates the intentional termination of the planning sequence. The planner identifies when the mission has been successfully completed or when no further actions are necessary, prompting the system to cease the planning loop efficiently.

3. \textbf{Impractical error signals} may cause hallucination planning or counterfactual execution. This mechanism is critical in preventing the execution of actions that could compromise system integrity or lead to task failure. These signals are divided into four subcategories:
\begin{itemize}
    \item Lack of Ability (LoA): The multi-robot system lacks the fundamental capabilities required for a task, often due to missing specific robot types, such as those with locomotion or manipulation abilities.
    \item Lack of Skill (LoS): A robot skill necessary for completing the task is unregistered in the robot configuration.
    \item Load Over Limit (LoL): The task requires manipulating objects that exceed the robot's load capacity .
    \item Lack of Object (LoO): Essential objects required for task execution are unavailable or absent in the environment description.
\end{itemize}

In summary, MultiPlan+ offers a diverse and meticulously annotated dataset that supports the fine-tuning of language models for next-action prediction in multi-robot systems. Its inclusion of error-handling mechanisms and diverse task scenarios makes it a critical resource for enhancing the generalizability and reliability of multi-robot planning.

\subsection{Robot Planning Assessment Schema (RPAS)}

To comprehensively evaluate robot planning sequences, we propose the Robot Planning Assessment Schema (RPAS), which incorporates multiple evaluation metrics to capture different aspects of planning performance. Based on the human-reviewed success rate and expert grading, the comprehensive metric RPAS is defined as \cref{eq:raps}

\begin{equation}
    RPAS = Avg(ASR_{top-k}, Expert) \times 100\%.
    \label{eq:raps}
\end{equation}

\subsubsection{Average Success Rate with Top-k Matching}

The top-k success rate metric is designed to evaluate how well a predicted planning sequence aligns with the reference sequence in the first $k$ steps. This is especially useful for assessing planning algorithms that may produce partially correct sequences, as it rewards agents for correctly planning the initial steps even if the later steps make mistakes. For predicted planning sequence $P$ and reference sequence $R$, the top-k average success rate is defined as \cref{eq:asr}.

\begin{equation}
    ASR_{top-k}(P, R) = \frac{1}{N}\sum_{i=1}^{N}\mathbf{1}\{\text{match}_k(P_i, R_i)\} \times 100\%,
    \label{eq:asr}
\end{equation}
where $N$ is the total number of samples, and $\text{match}_k(P_i, G_i)$ equals 1 if the $k$-th step of prediction matches the reference.

\subsubsection{Expert Grading with LLM}

To capture nuanced differences in planning quality, we employ a LLM-based grading system. This system leverages the advanced reasoning and contextual understanding capabilities of LLMs to evaluate the quality of planning sequences beyond simple exact matches. The LLM is prompted to assess the logical consistency, feasibility, efficiency, and robustness of the planning sequence. The LLM assigns a score on an overall scale of 0 to 100 with weighted criteria. This approach provides a more holistic evaluation of planning quality, capturing subtle differences that traditional metrics might miss.

\subsubsection{Multi-Robot Planning Error Diagnosis (MRED)}
Building on prior work \cite{multiplan}, we propose a structured error taxonomy to diagnose failures in multi-robot planning systems. Regarding the next action prediction inference paradigm, we add Ending Error (EE) to detect improper ending of the planning. This taxonomy enables granular root-cause analysis of planning failures, supporting targeted improvements in multi-robot systems. MRED categorizes errors as follows:

\begin{itemize}
    \item Unregistered Error (UE): Occurs when unregistered components are referenced. Sub-types include: UE\_robot: Use of an undeclared robot; UE\_skill: Reference to an undefined skill; UE\_obj: Deployment of an unregistered object; UE\_pos: Use of an unspecified spatial position.
    \item Position Error (PoE): Inaccuracies in spatial reasoning, such as ambiguous or conflicting position assignments.
    \item Planning Error (PlE): Generation of impractical or infeasible plans during robotic operations.
    \item Skill Error (SE): Incorrect parameterization of skill functions.
    \item Ending Error (EE): Premature halting before task completion, redundant continuation after goal fulfillment, and erroneous impractical error generation. 
\end{itemize}

\begin{table*}[ht]
    \centering
    \caption{\textbf{Performance of robot task planning with various core LLMs of EmbodiedAgent}, evaluated using RPAS metrics. Higher values indicate superior performance. Bold text highlights the best-performing model. The metrics represent the average percentages over 32 trials from the unseen test dataset, with results rounded.}
    \label{tab:comparison}
    \begin{tabular}{l|ccccccccc}
        \toprule
        \textbf{Model} & Infer type & \textbf{$ASR_{top-k}$} & Expert grading & UE & PoE & PlE & SE & EE & RPAS \\
        \midrule
        \multicolumn{10}{c}{\textbf{Proprietary}} \\
        \midrule
        GPT-4o & FPS & 9.03 & 31.25 & 0.00 & 0.00 & 6.25 & 93.75 & 0.00 & 20.14 \\
        OpenAI-o1 & FPS & 6.74 & 34.38 & 0.00 & 0.00 & 3.12 & 96.88 & 0.00 & 20.56 \\
        Claude-3.5-Sonnet & FPS & 37.5 & 32.19 & 0.00 & 0.00 & 0.00 & 0.00 & 62.5 & 34.85 \\
        \midrule
        \multicolumn{10}{c}{\textbf{Open-weight}} \\
        \midrule
        Deepseek-R1 & FPS & 61.87 & 64.84 & 0.00 & 18.75 & 15.62 & 0.00 & 18.75 & 63.36 \\
        Distill-Llama3.3-70B\tablefootnote{Distill by Deepseek-R1} & NAP & 33.70 & 60.00 & 3.12 & 59.38 & 3.12 & 0.00 & 21.88 & 46.85\\
        LLaMA-3.1-8B & NAP & 16.56 & 43.44 & 3.12 & 71.88 & 50.00 & 18.75 & 3.12 & 30.00 \\
        LLaMA-3.1-70B & NAP & 27.50 & 40.94 & 0.00 & 9.38 & 87.5 & 62.5 & 21.88 & 34.22 \\
        LLaMA-3.1-405B & NAP & 34.24 & 61.09 & 28.12 & 46.88 & 6.25 & 0.00 & 0.00 & 47.67 \\
        LLaMA-3.3-70B & NAP & 46.02 & 60.31 & 0.00 & 48.48 & 39.39 & 15.15 & 6.06 & 53.26\\
        Gemma-2-9B & NAP & 17.36 & 29.06 & 31.25 & 34.38 & 25.0 & 0.00 & 6.25 & 23.21 \\
        Qwen2.5-7B & NAP & 9.17 & 10.94 & 3.12 & 15.62 & 90.62 & 0.00 & 6.25 & 10.06 \\
        Qwen2.5-72B & NAP & 40.42 & 50.62 & 0.00 & 56.25 & 18.75 & 0.00 & 6.25 & 45.52 \\
        MAP-Neo-7B-Multiplan & FPS & 46.39 & 44.22 & 21.88 & 28.12 & 25.0 & 6.25 & 3.12 & 45.31 \\
        \textbf{EmbodiedAgent (Ours)} & NAP & \textbf{74.01} & \textbf{69.69} & 9.38 & 15.62 & 3.12 & 9.38 & 0.00 & \textbf{71.85} \\
        \bottomrule
    \end{tabular}
\end{table*}

\begin{figure}
    \centering
    \includegraphics[width=1\linewidth]{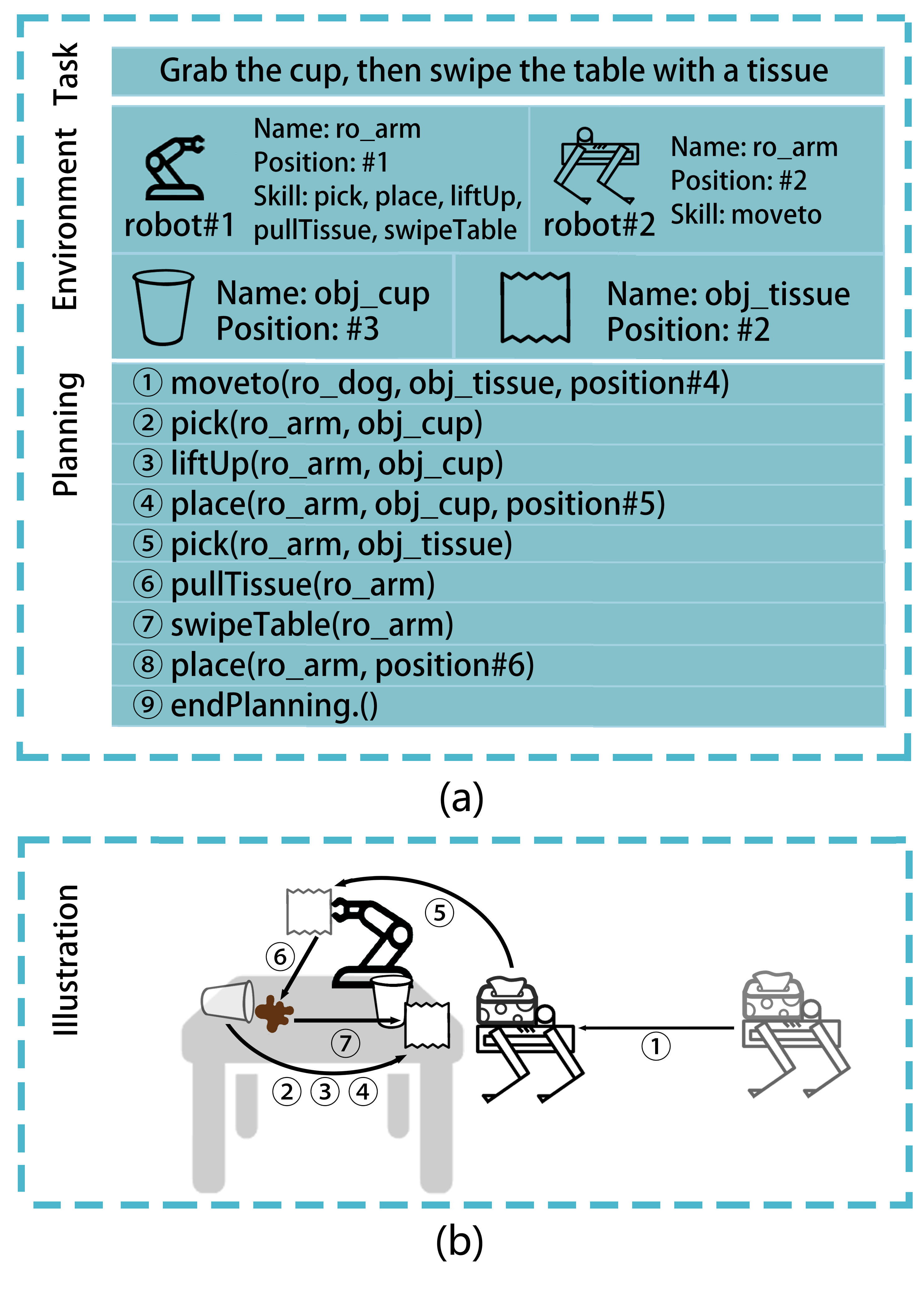}
    \caption{This figure introduces the planning task for real-world experiment. (a) outlines the mission specification, environment description and planning sequence. (b) visually illustrates the task execution, highlighting the robots' movements and object interactions during the task.}
    \label{fig:real-ill}
\end{figure}

\section{EXPERIMENTS}
The proposed method EmbodiedAgent is validated through comparisons with various proprietary and open-source LLMs in both next-action prediction and full action sequence prediction settings. Additionally, a real-world experiment is conducted to assess the deployment of EmbodiedAgent in an office service scenario.

\textbf{Experiment Setups and Dataset} The Llama-3.1-8B-Instruct \cite{llama} model is fine-tuned using the proposed MultiPlan+ dataset in a next-action prediction framework. The supervised fine-tuning process is carried out on a high-performance computational cluster equipped with eight NVIDIA A100 GPUs (80GB), running for three epochs over approximately eight hours with a learning rate of 2e-4 and a batch size of 32. 

The test dataset contains 32 unseen tasks in MultiPlan+ with 2 impractical data samples. All the experiments are conducted in a one-shot inference manner. Model evaluation is conducted using the RPAS metrics, ensuring a comprehensive assessment of its performance. 

To further validate the proposed method, a real-world experiment is conducted in an office service scenario, where a robotic arm collaborates with a quadrupedal robot to complete a structured task. Specifically, the robotic arm is responsible for lifting a cup and wiping spilled coffee using tissues carried by the quadrupedal robot. The low-level control of the robotic arm was implemented using the Action Chunk Transformer (ACT) \cite{act}, a transformer-based model designed for robotic manipulation, which mitigates compounding errors in traditional imitation learning approaches through chunked actions and temporal ensemble techniques. Concurrently, the quadrupedal robot was controlled via an integrated software development kit (SDK), utilizing its built-in locomotion capabilities to navigate and perform errand services. This experiment not only showcases the effectiveness of EmbodiedAgent in orchestrating multi-robot collaboration for complex service tasks but also underscores the practical applications of heterogeneous robot teams in office environments, where autonomous systems can assist with daily tasks efficiently.

\subsection{Heterogeneous Multi-Robot Task Planning}  

To comprehensively evaluate \textit{EmbodiedAgent}, we benchmark its task planning performance against state-of-the-art proprietary and open-source large language models (LLMs) across two key metrics: next-action prediction (NAP) and full planning sequence (FPS) generation. Our evaluation is conducted within the \textbf{RPAS framework}, which incorporates multiple dimensions of task execution quality, including the top-k action success rate (ASR\textsubscript{top-k}), expert grading scores, Multi-Robot Error Diagnosis (MRED) categories, and the overall composite score RPAS, ensuring a rigorous and multi-faceted assessment of planning efficacy.

As shown in Table \cref{tab:comparison}, \textit{EmbodiedAgent} establishes a new state-of-the-art in multi-robot task planning, achieving an impressive \textbf{74.01\% ASR\textsubscript{top-k}} and \textbf{71.85\% RPAS}, surpassing all other benchmarked models. A deeper comparative analysis highlights critical shortcomings in models such as GPT-4o and OpenAI-o1, where early-stage skill errors propagate through the planning sequence, leading to fundamental breakdowns in plan validity and execution feasibility. Notably, results emphasize the importance of task-specific optimization: despite its relatively compact 7-billion-parameter configuration, the fine-tuned MAP-Neo-7B-Multiplan model delivers highly competitive RPAS performance, significantly outperforming larger-scale models such as LLaMA-3.1-70B, particularly in structured multi-robot planning scenarios. This underscores the effectiveness of domain-adaptive training over naive parameter scaling, reinforcing the necessity of specialized model fine-tuning for embodied AI applications in complex, real-world robotics tasks.

\footnotetext[2]{Distilled by Deepseek-R1}

\subsection{Multi-Robot System for Office Service}

\begin{figure*}
    \centering
    \includegraphics[width=0.95\linewidth]{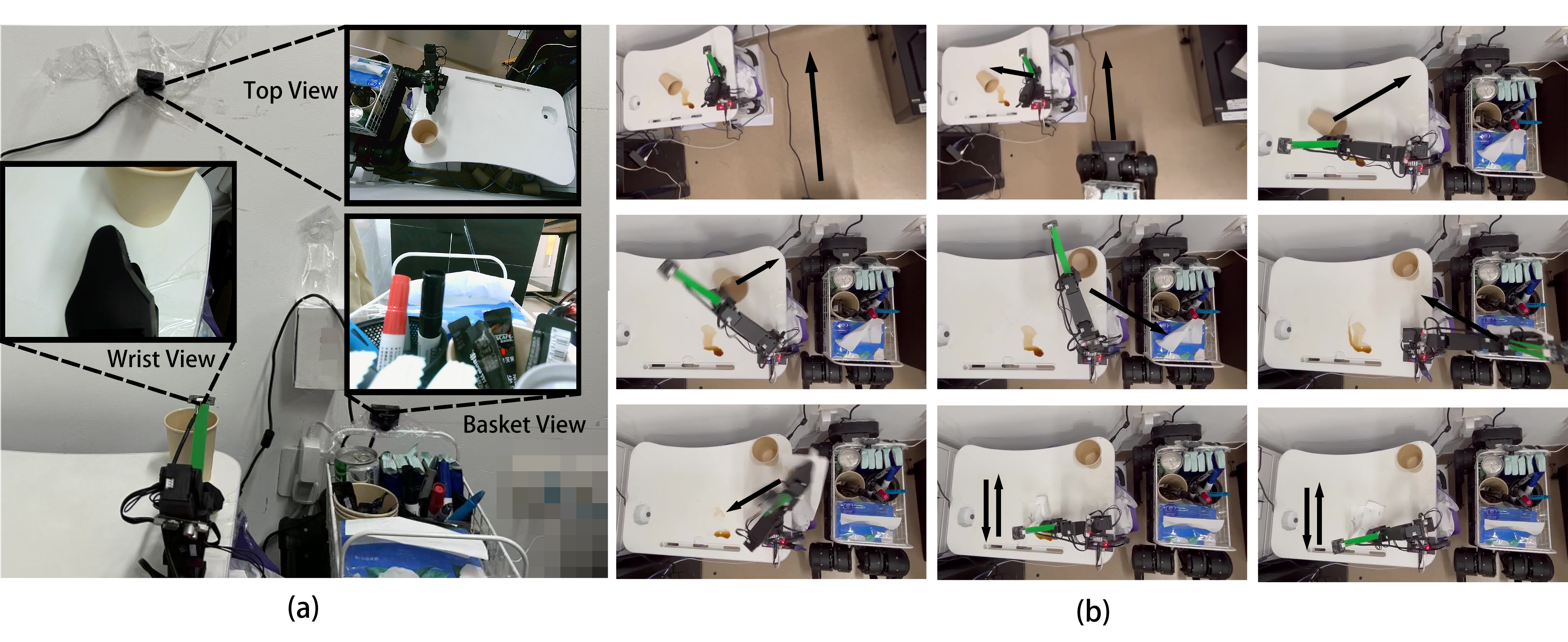}
    \caption{(a) Overview of the office service experiment. Showcasing the position and corresponding frames different views: wrist view, top view, and basket view. (b) Captured frames during the office service experiment. A robot dog delivers tissue for the robot arm to clean the spilled coffee. The robot arm first lifts the cup and then swipes the table. Arrows denote the robot's intended movement.}
    \label{fig:real}
\end{figure*}

To validate the robustness of EmbodiedAgent in real-world settings, we designed an office service scenario for testing. The mission description, environment setups, planning sequence, and a conceptual illustration are shown in \cref{fig:real-ill}. For low-level execution, we employ the Action Chunking Transformer (ACT) to train the robot arm on individual tasks and utilize the SDK for the robot dog. To enable autonomous task execution by the robot arm, we first collected demonstration data through teleoperation, where human operators remotely controlled the arm. A total of 50 demonstration episodes were collected for each task, with each episode lasting approximately 15-20 seconds. Video data was captured from three camera views: a wrist camera view, a top-down view, and a view of the basket mounted on the back of the robot dog, as shown in \cref{fig:real}a. This experiment assessed the effectiveness of the robot arm and robot dog system in performing complex, captured frames are shown in \cref{fig:real}b. It evaluated the system's ability to transfer learned behaviors to new, unseen office scenarios, focusing on adaptability, efficiency, and overall performance under real-world conditions.

\section{CONCLUSIONS}
EmbodiedAgent represents a significant advancement in multi-robot planning by seamlessly integrating hierarchical control. The MultiPlan+ dataset, structured in a next-action prediction format, is further augmented with impractical tasks to enhance robustness. Through supervised fine-tuning on MultiPlan+, the planner effectively addressing the issue of impractical actions in dynamic environments. The MultiPlan+ dataset and RPAS metrics introduced in this work serve as essential tools for benchmarking and evaluating future developments in the field of multi-robot systems. Future works may explore the capabilities of EmbodiedAgent through more applications in the real-world.

\addtolength{\textheight}{-12cm}   



\end{document}